\documentclass{article}
\usepackage{spconf,amsmath,graphicx, tikz, amsfonts}

\title{Investigating SINDy As a Tool For Causal Discovery In Time Series Signals}
\name{Andrew O'Brien, Rosina Weber, Edward Kim}
\address{Drexel University }
\begin{document}
%
\maketitle
\begin{abstract}
The SINDy algorithm has been successfully used to identify the governing equations of dynamical systems from time series data.  In this paper, we argue that this makes SINDy a potentially useful tool for causal discovery and that existing tools for causal discovery can be used to dramatically improve the performance of SINDy as tool for robust sparse modeling and system identification. We then demonstrate empirically that augmenting the SINDy algorithm with tools from causal discovery can provides engineers with a tool for learning causally robust governing equations. 

\end{abstract}
\section{Introduction}
\label{sec:intro}

The Sparse Identification of Nonlinear Dynamics (SINDy) algorithm was proposed by  Brunton et al. \cite{Brunton:2016} as a tool for learning the governing equations of  a dynamic system from data collected from the system as it evolved through time. It has been used extensively in the engineering and signal analyses communities  to recover governing  equations  in a diverse set of circumstances\cite{Champion:2020} \cite{Mangan:2016} \cite{Rudy:2017} .

	Within the artificial intelligence community, a related set of problems related to causality have recently generated considerable interest \cite{Jakob:2019} \cite{Spirtes:2016} \cite{Kaheman:2022}. The ability to take as input a data set and return as output the causal relations between variables in that data set is a central challenge known as causal discovery. A wide variety of algorithms are currently in use within the field \cite{Heinze-Deml:2018}. 
	
	Peters et al. \cite{Peters:2020} formalized the exact connection between learning a set of ordinary differential equations  that govern a system and identifying the causal relations between the variables within a system . In their framework, a variable x can be said to be a direct cause of a variable y if and only if x is present in the ordinary differential equation governing the change in y. Therefore, an algorithm can successfully learn the ordinary differential equations governing a system only if it is also learning the causal relations within a system. This makes SINDy a potentially valuable and, to the best of our knowledge, underexplored tool for causality researchers. The connection should also open the door to scientists, engineers, and signal process researchers to use the recently developed tools of causal discovery to improve the performance of SINDy as a system identification tool. 
	
In this paper we compare SINDy’s ability to identify causal relationships between variables against a sampling of prominent causal discovery algorithms. We first show that under the assumption that all measured variables are part of the same dynamical system, SINDy achieves similar performance to the comparison algorithms. Once this assumption is relaxed, we show the SINDy’s performance degrades substantially. Finally, we show that augmenting SINDy with a comparison algorithm can make SINDy a state-of-the-art tool for causal discovery and a substantially more robust tool for engineers and signal process researchers performing sparse modeling and system identification.

\section{Background}
\label{sec:background}

\subsection{Sparse Regression}
\label{ssec:sparse}

Regression is a branch of machine learning that takes some data set $\{(X_{i}, Y_{i})\}_{i=1}^{N}$  and some parameterized function $f(\cdot,\beta)$ and attempts to use the data set to find a set of parameters $\beta^{*}$ such that $f(X_{i};\beta^{*}) \approx Y_{i}$. Typically $X_{i},\beta  \in \mathbb{R}^{k x 1}$ and $Y_{i} \in \mathbb{R}$. A common approach to regression is assuming f is linear and trying to find the set of parameters that minimize the $l_{2}$ norm of the difference vector between the model predictions and the ground truth. This approach is shown in Equation 1 where the rows of $X$ are the transpose of the $X_{i}$ data vectors and the elements of Y are the $Y_{i}$ scalars. 

\begin{equation}
\beta^{*} = \underset{\beta }{\mathrm{argmin}} (|| X^{T}\beta - Y||_{2})
\end{equation}

There are common problems with this approach. If the regression is under-determined, there are often multiple solutions to Equation 1.  In the high-dimensional case, $\beta$ is often hard to interpret for stakeholders making use of the model solutions. Finally, solutions with large amounts of parameters frequently don't generalize well to out-of-sample data. All of these problems can be solve by imposing an $l_{0}$ sparseness penalty to the optimization problem in Equation 1. In practice, it is difficult to perform this optimization directly so the $l_{1}$ norm is used as a relaxation of the $l_{0}$ norm as depicted in Equation 2.

\begin{equation}
\beta^{*} = \underset{\beta }{\mathrm{argmin}} (||X\beta - Y||_{2} + \lambda ||\beta||_{1})
\end{equation}

\begin{figure}
\includegraphics[width=7cm, height=7cm,]{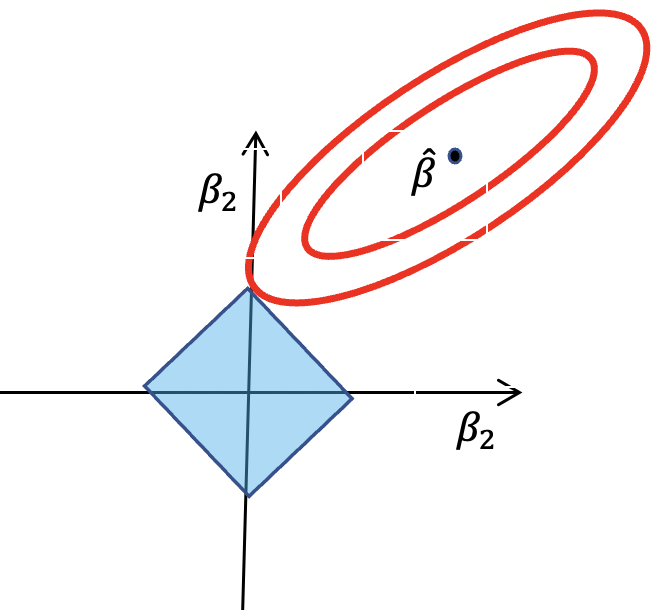}
\caption{2-d example of $l_{1}$ regression's tendency  to promote sparse solutions. This illustration is adapted from an image in \cite{ Tibshirani:1996}.}
\end{figure}

\subsection{SINDy}
\label{ssec:sindy}

The SINDy algorithm takes as input time series data and attempts to reconstruct the dynamical system that generated the data \cite{Brunton:2016}. An autonomous dynamical system is formally defined as $\frac{d}{dt}\textbf{x}(t) = \textbf{f}(\textbf{x}(t); \boldsymbol \beta )$ where t is the time,  $\textbf{x}(t)$ is the state of the system at time t, $\boldsymbol \beta$ is a set of system parameters, and $\textbf{f}$ is a governing function describing how the system state is changing at time t. Governing functions of dynamic systems are often sums of a few simple functions \cite{Silva:2020}. SINDy treats learning theses governing functions as a sparse regression problem. 

Formally, the algorithm takes as input two matrices $\textbf{X} = [x(t_1), ..., x(t_m)]^{T}$ and  $\boldsymbol{ \dot{X} } = [\dot{x}(t_1), ..., \dot{x}(t_m)]^{T}$ where $x(t_1) \in \mathbb{R}^{n x 1}$ is the measurement of the state of the system at time t, $\dot{x}(t_1) \in \mathbb{R}^{n x 1}$, is the measurement (or approximation) of the derivative of the system at time t. 
From $\textbf{X}$, a library of candidate functions is constructed and denoted by $\boldsymbol \theta (\textbf{X})$. The optimization in Equation 3 is then performed for each of the N columns of $\boldsymbol{\dot{X}}$.

\begin{equation}
\xi_{k} = \underset{\xi_{k}^{'} }{\mathrm{argmin}} (\vert \vert \textbf{\.X}_{k} - \boldsymbol \theta (\textbf{X}) \xi_{k}^{'} \vert \vert_{2} + \lambda \vert \vert \xi_{k}^{'} \vert \vert_{1})
\end{equation}

We solve the optimization problem in Equation 3 using the sequential thresholding least squares (STLS) proposed in \cite{Brunton:2016}. The STLS algorithm works by first performing a standard least squares regression of $\textbf{\.X}$ on $\theta$ to get initial values for $\xi_{k}$. A threshold c is then chosen such that any element of $\xi_{k}$ less than c is set to 0. Then a least squares regression is again performed of $\textbf{\.X}$ onto the columns of $\theta(X)$ that have non-zero coefficients in $\xi_{k}$. The process of least squares regression and thresholding is continued until the values of $\xi_{k}$ converge

\subsection{Causality}
\label{ssec:causality }
The dominant causal framework in artificial intelligence consists of a structural causal model (SCM) M and a corresponding directed causal graph $G_{M}$ \cite{Pearl:2009}. M = (U, V, F) where V is a set of endogenous variables, U is a set of exogenous variables, and F  is a set of assignment functions that assign values to the variables in V using the values of the variables in V and U. A probability distribution over U induces a probability distribution over V as well. For every element of U and every element of V, there is a node in $G_{M}$. There is a directed edge from a node $n_{i}$ to a node  $n_{j}$ if and only if variable i is an argument of variable j’s assignment function $f_{j}$. We say that i is a direct cause of j if and only if i is a parent of j in graph $G_{M}$. 

For illustrative purpose, consider the following simple example. Imagine some system with variables X,Y,Z,$U_{X}$, and $U_{Y}$. $U_{X}$ and $U_{Y}$ are exogenous variables with standard normal probability distributions. X's value is a function of $U_{X}$.Y is a linear function of X and $U_{Y}$. Z is a liner function of Y and X. The following is a potential SCM, $M = (U, V, F )$ where  $V= \{X, Y, Z\}$, $U = \{U_{X}, U_{Y}\}$, and $F = \{f_{X}, f_{Y}, f_{Z} \}$ such that 

\begin{equation} 
\begin{split}
X := f_{X}(U_{X}) = 10 * U_{X} + 3, U_{X} \sim N(0,1)  \\
Y := f_{Y}(X, U_{Y}) = 2X + U_{Y}, U_{Y} \sim N(0,1) \\
Z := f_{Z}(X, Y) = 5X + 4Y\\
\end{split}
\end{equation}

The corresponding causal graph $G_{M}$ is given in Figure 2.

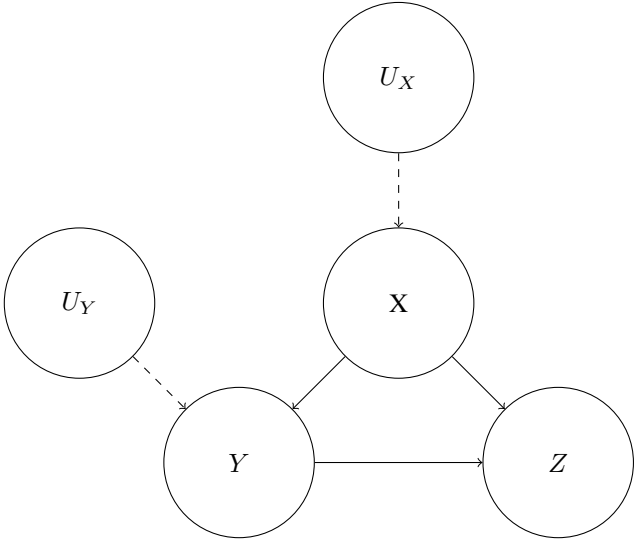
\begin{figure}
\begin{center}

\begin{tikzpicture}[node distance={30mm},  main/.style = {draw, circle, minimum size=2cm}] 
\node[main] (4) {$U_{X}$};
\node[main] (1)  [below of =4] {X};
\node[main] (2) [below left of=1] {$Y$};
\node[main] (5)  [above left of=2] {$U_{Y}$} ;
\draw[->] (1) -- (2);
\node[main] (3) [below right of=1] {$Z$};
\draw[->] (2) -- (3);
\draw[->] (1) -- (3);
\draw[dashed,->] (5) -- (2);

\draw[dashed,->] (4) -- (1);
\end{tikzpicture} 
\caption{The causal graph corresponding to the structure equations in Equation 4. Solid lines indicate causal relations between endogenous variables and dashed lines indicate causal relations between an exogenous and an endogenous variable.}

\end{center} 
\end{figure}

The causal framework can be extended to dynamical systems \cite{Peters:2020}. Consider a set of d ordinary differential equations and initial values assignments. These can be written in as assignment equations with the form given in Equation 5. The causal graph is defined analogously for the dynamical case. This causal interpretation of dynamical systems allows for a causal interpretation of the ordinary differential equations learned by the SINDy algorithm and thus to construct a casual graph from them. 

Causal discovery is the process of taking data generated from the distribution defined by a SCM and inferring either the model's causal graph or the structural equations of that model \cite{Spirtes:2016}. The problem is difficult because the SCM is underdetermined by the data. Many causal identification algorithms attempt to solve this problem by making assumptions about the underlying SCM or causal graph and the algorithms can be classified according to the kind of assumption they make and how much they are trying to infer \cite{Heinze-Deml:2018}. In causal identification with time series data, methods often assume either that conditional independence in the data implies conditional independence in the causal graph, the structural equations in the SCM have a particular parametric form and the exogenous variables have a particular probability distribution, or assume that the structural equations are a dynamical system with a reconstructable state space, \cite{Jakob:2019} \cite{Moraffah:2021}.

\begin{equation}
\frac{d}{dt} x_{t}^{i} := f^{i}(x_{t}^{PA_{i}}), x_{0}^{i} := \alpha_{0}^{i}
\end{equation}

\section{Methodology}
\label{sec:print}

 Two sets of experiments were run to test SINDy's ability as an algorithm for causal identification. A final experiment was run to test the ability of existing causal identification techniques to improve SINDy's performance.  For the first set of experiments, a diverse set of 6 dynamical systems were chosen which are described in table 1. For each dynamical system, 10 simulations were run. For each simulation, the Runge-Kutta algorithm \cite{Teukolsky:2007} was used to simulate the system for 1000 time steps. The SINDy algorithm and 4 causal identification algorithms were used to construct the causal graph of the system from the data. The 4 comparison causal identification algorithms  were chosen on the basis of diversity of technique, breadth of use, and performance (Section 3.1). 
 
The ground truth causal graph was constructed for the dynamical systems using the framework described in \cite{Peters:2020}. The ground truth causal graph and learned causal graphs were represented as adjacency matrices. The Hamming distance between the learned adjacency matrix and the ground truth matrix was used as the metric to measure the quality of the learned causal graphs. The Hamming distance between the ground truth causal graph and the learned graph is computed for each algorithm after each simulation of the dynamical system. The average Hamming distance for all 10 simulations is computed for each of the six dynamical systems. 

The second set of experiments are similar except Gaussian noise variables are added to double the size of the system. This represents the case in which the scientist using these techniques is uncertain as to which variables are part of the system. 

For the third set of experiments, the PCMCI algorithm was first run as a prepossessing step. Missing edges in the the causal graph learned by PCMCI were then encoded as a set of constraints the SINDy algorithm had to obey. The new constraint augmented SINDy algorithm was run on the same set of simulated dynamical systems as in experiment 1 and 2 
 
\begin{table}[h!]

\resizebox{\columnwidth}{!}{%

\begin{tabular}{ | c | c|  } 

 \hline
 System & System Equations \\
 \hline
 Lorenz \cite{Lorenz:1963} & $\begin{array} {lcl} \\ \dot{x} & = & 10(y-x) \\ \\ \dot{y} & = & x(28 - z)\\ \\
 \dot{z} & = & xy - \frac{8}{3}z  \end{array}$ \\ 
 \hline
 Mankiw-Romer-Weil \cite{Mankiw:1992} & $\begin{array} {lcl} \\ \dot{k} & = & .01(kh)^\frac{1}{3} - .06k \\ \\ \dot{h} & = & .01(kh)^\frac{1}{3} - .06h 
\end{array}$   \\ 
 \hline
 FitzHugh-Nagumo \cite{Izhikevich:2010} & $\begin{array} {lcl} \\ \dot{v} & = & v(.1-v)(v-1) - w + 5 \\ \\ \dot{w} & = & .01v - .02w
\end{array}$  \\ 
 \hline
Lotka-Volterra \cite{Vano:2006} & $\begin{array} {lcl} \\ \dot{n_{1}} & = & n_{1}(1-n_{1} - 1.09n_{2} - 1.52n_{3}) \\ \\ \dot{n_{2}} & = & .72n_{2}(1-n_{2} - .44n_{3} - 1.36n_{4}) \\ \\ \dot{n_{3}} & = & 1.53n_{3}(1-2.33n_{1} - n_{3} - .47n_{4}) \\ \\ \dot{n_{4}} & = & 1.27n_{4}(1-1.21n_{1} - .52n_{2} - 1.53n_{3} - n_{4})
\end{array}$ \\ 
 \hline
Pendulum \cite{Strogatz:2015} & $\begin{array} {lcl} \\ \dot{u} & = & v \\ \\ \dot{v} & = & -.76sin(u)
\end{array}$  \\
 \hline
SIR\cite{Thurner:2018} & $\begin{array} {lcl} \\ \dot{s} & = & -\frac{.715}{60}is \\ \\ \dot{i} & = & \frac{.715}{60}is - .285i \\ \\
\dot{r} & = & .285i \\
\end{array}$ \\ \\
 \hline
\end{tabular}

}
\caption{Dynamical systems used to test SINDy as a tool for causal inference.}
\label{table:1}
\end{table}

\subsection{Comparison Algorithms}
\label{ssec:comparison}

The Convergent Cross Mapping (CCM) algorithm \cite{Sugihara:2012} starts by using Takens's theorem \cite{Takens:1980} as a justification to reconstruct the system manifold, M,  twice using time lags from two system variables x(t) and y(t). The reconstructions using x(t) lags and y(t) lags are denoted $M_{x}$ and $M_{y}$ respectively. If the ability to predict $M_{y}(t)$ from points on $M_{x}(t)$ in some neighborhood around t increases as the number of sampled points increases, then y(t) is a cause of x(t). 

The premise behind the Granger Causality (GC) algorithm \cite{Granger:1969} is that a variable x(t) is a direct cause of y(t) if and only if the past values of x(t) contain unique information about y(t). In practice, bivariate Granger Causality is used whereby if a linear regressive model using the time lags of y(t) and x(t) can produce better predictive value than just the lags of y(t) then x(t) is said to be a cause of y(t). 

The PCMCI algorithm \cite{Runge:2019} is an extension of the PC algorithm \cite{Spirtes:2001}. The PC algorithm assumes that conditional independence in the data implies conditional independence in the causal graph. This assumption is called causal faithfulness. The central idea of the algorithm is that, under the assumption of causal faithfulness, variables that are conditionally independent in the data set have no edge connecting them in the graph. Therefore repeated conditional independence tests can be used to "prune" a complete directed graph.

The Linear Nongaussian Acyclic Model \cite{Shimizu:2006} assumes the assignment functions in the SCM are linear and that the exogenous variables have a non-Gaussian probability distribution. Under these assumptions, it can be proven that there is a unique SCM that corresponds to the probability distribution that generated the data and it can be discovered by using independent component analysis. 

\section{Results \& Conclusion}
\label{sec:page}

The results from experiment 1 are described in Table 2. On average, SINDy identified the correct graph as well or better than 2.67 of the other 4 techniques techniques. It had the best average performance in 2 of the 6 systems. This is evidence that SINDy is a strong tool for causal identification in dynamical systems under the assumptions that all system variables and only system variables have been measured. 

In experiment 2 (Table 3), SINDy performed much worse relative to the other algorithms. It  achieved the best average performance on 0 systems and on average only performed as well .67 comparison algorithms per dynamical system. To the best of our knowledge, the degradation of SINDy's performance in the presence of non-system variables is a weakness of the algorithm that is currently absent from the literature. It also suggests that engineering and signal processing researchers could greatly improve the performance of SINDy by first using established causal identification techniques as a pruning step when unsure of the exact number of variables in the system. Experiment 3 tested this and demonstrated that the deficiency could be remedied using PCMCI as a preprocesssing step. When preprocessing was done, (augmented) SINDy achieved best performance on 4 of the 6 systems and on average outperformed 3.67 of the 4 other methods. 

This state-of-the art performance on causal identification provides strong evidence that using SINDy with a preprocessing step is a potentially very useful technique for artificial intelligence researchers in causal identification and engineers in system identification. 

This study used numerical experiments only. In the future, data from real dynamic systems should be used.

\begin{table}[h!]

\resizebox{\columnwidth}{!}{%

\begin{tabular}{ |c|c|c|c|c|c|c|  } 

 \hline
 & PCMCI \cite{Runge:2019} & LINGAM \cite{Shimizu:2006} & GC \cite{Granger:1969} & CCM \cite{Sugihara:2012} & SINDy \cite{Brunton:2016} \\
 \hline
 Lorenz & .11 & .34 & .11 & .11 & .01 \\
 \hline
 MRW  & .65 & .70 & .00 & .35 & .40 \\ 
 \hline
 FN  & .00 & .33 & .00 & .00 & .03 \\ 
 \hline
 LV  & .24 & .41 & .19 & .19 & .22 \\
 \hline
  Pendulum  & .50 & .88 & .50 & .50 & .43    \\
 \hline
   SIR  & .49 & .40 & .44 & .44 & .43 \\
 \hline

\end{tabular}

}
\caption{Hamming loss for SINDy and comparison methods for each system}
\label{table:3}
\end{table}

\begin{table}[h!]

\resizebox{\columnwidth}{!}{%

\begin{tabular}{ |c|c|c|c|c|c|c|  } 

 \hline
 & PCMCI \cite{Runge:2019} & LINGAM \cite{Shimizu:2006} & GC \cite{Granger:1969} & CCM\cite{Sugihara:2012} & SINDy \cite{Brunton:2016} \\
 \hline
 Lorenz & .06  & .43  & .35  & .65  & .64  \\
 \hline
 MRW  & .19  & .44  & .43  & .57  & .73  \\ 
 \hline
 FN  & .05  & .52  & .31  & .65  & .75  \\ 
 \hline
 LV  & .11 & .51 & .34 & .67 & .80  \\
 \hline
  Pendulum  & .21 & .68 & .43 & .75 & .88    \\
 \hline
   SIR  & .62  & .68 & .44 & .35 & .42 \\
 \hline

\end{tabular}

}
\caption{Hamming loss for SINDy and comparison methods for each system with noise variables}
\label{table:4}
\end{table}

\begin{table}[h!]
\begin{center}

\begin{tabular}{ |c|c|  } 

 \hline
 System & Augmented SINDy\\
 \hline
 Lorenz & .07\\
 \hline
 MRW  & .13  \\ 
 \hline
 FN  & .14\\ 
 \hline
 LV & .11 \\
 \hline
  Pendulum  & .17 \\
 \hline
   SIR  & .15\\
 \hline

\end{tabular}

\caption{Hamming loss for SINDy after PCMCI was run as a pre-pruning step}
\label{table:5}
\end{center}
\end{table}

\bibliographystyle{IEEEbib}
\bibliography{refs}

\end{document}